\documentclass[11pt]{article}

\usepackage[preprint]{acl}

\usepackage{times}

\usepackage[T1]{fontenc}

\usepackage[utf8]{inputenc}

\usepackage{microtype}

\usepackage{inconsolata}

\usepackage{graphicx}
\usepackage{algorithm}
\usepackage{algorithmic}
\usepackage[dvipsnames]{xcolor}
\usepackage{amsmath}
\usepackage{booktabs}
\usepackage{multirow}
\usepackage{subcaption}
\usepackage{listings}
\definecolor{promptkeyword}{named}{RoyalBlue}
\definecolor{promptplaceholder}{named}{PineGreen}

\title{LLM-MemCluster: Empowering Large Language Models with Dynamic Memory for Text Clustering}

\author{
  \textbf{Yuanjie Zhu}\textsuperscript{\rm 1}\quad
  \textbf{Liangwei Yang}\textsuperscript{\rm 1}\quad
  \textbf{Ke Xu}\textsuperscript{\rm 1}\quad
  \textbf{Weizhi Zhang}\textsuperscript{\rm 1}\\
  \textbf{Zihe Song}\textsuperscript{\rm 1}\quad
  \textbf{Jindong Wang}\textsuperscript{\rm 2}\quad
  \textbf{Philip S. Yu}\textsuperscript{\rm 1} \\
  \textsuperscript{1}University of Illinois Chicago \\
  \textsuperscript{2}William \& Mary \\
  \texttt{\{yzhu224, lyang84, kxu25, wzhan42, zsong29, psyu\}@uic.edu} \\
  \texttt{jwang80@wm.edu}
}

\begin{document}
\maketitle

\begin{abstract}
    Large Language Models (LLMs) are reshaping unsupervised learning by offering an unprecedented ability to perform text clustering based on their deep semantic understanding. However, their direct application is fundamentally limited by a lack of stateful memory for iterative refinement and the difficulty of managing cluster granularity. As a result, existing methods often rely on complex pipelines with external modules, sacrificing a truly end-to-end approach. We introduce \textbf{LLM-MemCluster}, a novel framework that reconceptualizes clustering as a fully LLM-native task. It leverages a \textbf{Dynamic Memory} to instill state awareness and a \textbf{Dual-Prompt Strategy} to enable the model to reason about and determine the number of clusters. Evaluated on several benchmark datasets, our tuning-free framework significantly and consistently outperforms strong baselines. LLM-MemCluster presents an effective, interpretable, and truly end-to-end paradigm for LLM-based text clustering.
\end{abstract}
\section{Introduction}

Text clustering, a cornerstone task in Natural Language Processing (NLP), aims to automatically organize a collection of documents into meaningful groups based on content similarity. This unsupervised learning technique is pivotal for large-scale knowledge discovery and information organization, with its utility demonstrated in applications ranging from structuring massive document archives to analyzing the collective voice of online communities~\cite{10.1145/3689036, hadifar-etal-2019-self}. Traditional clustering methods, such as K-Means~\cite{jin2017k, sinaga2020unsupervised} or hierarchical clustering~\cite{sahoo2006incremental, ran2023comprehensive}, typically operate on vector-space representations like TF-IDF~\cite{7754750} or, more recently, pre-trained text embeddings from benchmarks like MTEB~\cite{muennighoff-etal-2023-mteb}. While these approaches are effective, a notable limitation~\cite{ezugwu2022comprehensive} is their reliance on either handcrafted features or domain-specific fine-tuning to achieve optimal performance.

The advent of Large Language Models (LLMs) with powerful semantic understanding and reasoning capabilities, such as GPT-4, Gemini, and DeepSeek~\cite{achiam2023gpt, team2023gemini, liu2024deepseek}, has introduced a new paradigm for text clustering. Current research, however, has largely focused on hybrid frameworks that employ LLMs in auxiliary roles to enhance traditional embedding-based pipelines. These applications include enriching text representations~\cite{wang-etal-2024-improving-text}, refining cluster assignments~\cite{feng-etal-2024-llmedgerefine}, and supervising the fine-tuning of external embedding models~\cite{zhang-etal-2023-clusterllm}. While innovative, these methods' reliance on external components precludes fully LLM-native clustering.

However, using LLMs as standalone clustering agents reveals two fundamental architectural challenges. The first is a direct conflict between operational requirements and model design: the limited context window necessitates processing large datasets in batches, yet the models' inherent statelessness prevents memory retention across these batches. This contradiction is a primary hurdle for achieving coherent and stable cluster assignments. This problem is further compounded by a second critical challenge: controlling clustering granularity. Without an explicit mechanism for guiding the partitioning process, LLMs tend to produce arbitrary and unstable topic partitions, as they lack an intrinsic method to determine a suitable degree of specificity. These limitations highlight the need for a framework that can impose statefulness while actively steering the clustering process.

To address these challenges, we introduce a novel framework for text clustering named \textbf{LLM-MemCluster}. This approach leverages large language models, requires no model fine-tuning or integration with traditional algorithms, and is driven by two key innovations---each specifically designed to address the aforementioned limitations.
\begin{enumerate}
    \item \textbf{Dynamic Memory Mechanism:} We introduce a memory mechanism that maintains a dynamic set of cluster labels within the prompt. This evolving memory state transforms the LLM into a state-aware clustering agent that can iteratively assign documents to existing clusters, create new ones for distinct topics, and merge and refine the cluster labels to ensure global consistency.
    \item \textbf{Granularity Control Mechanism:} To actively guide the LLM in determining a suitable number of clusters, we employ two distinct prompting modes. A strict prompt encourages the consolidation of the existing cluster memory into broader categories, whereas a relaxed prompt fosters the discovery of more fine-grained topics. This dual-mode strategy enables the framework to explore different levels of granularity, ultimately achieving a stable and well-justified cluster count.
\end{enumerate}

Our comprehensive experiments on several public benchmark datasets demonstrate that LLM-MemCluster significantly outperforms both traditional embedding-based methods and existing LLM-enhanced baselines across multiple standard evaluation metrics. These findings validate our framework as an effective solution for text clustering, harnessing the potential of end-to-end LLMs.

In summary, our contributions are threefold:
\begin{itemize}
    \item Dynamic Memory Mechanism overcoming inherent LLM statelessness and facilitating the iterative refinement of cluster quality.
    \item Granularity Control Mechanism employing a novel dual-prompt strategy to ensure precise, user-guided control of cluster granularity.
    \item State-of-the-art performance on multiple standard clustering benchmarks, demonstrating robust, fine-tuning-free generalization across a diverse spectrum of both proprietary and open-source large language models.
\end{itemize}

\section{Method}

\subsection{Problem Formulation}

Text clustering aims to automatically organize a collection of documents into meaningful groups based on content similarity. Formally, given an unlabeled text corpus, $\mathcal{D} = \{x_1, x_2, \dots, x_N\}$, the objective is to derive a partition of the corpus, $\mathcal{C} = \{\mathcal{C}_1, \mathcal{C}_2, \dots, \mathcal{C}_K\}$. This partition consists of $K$ clusters, where each cluster $\mathcal{C}_k$ is a subset of the original corpus $\mathcal{D}$, formally defined as:
$$ \mathcal{C}_k = \{x_j \in \mathcal{D} \mid l_j = k\} $$
Here, $l_j$ represents the label assigned to instance $x_j$. The final partition covers all instances, with mutually exclusive clusters. The number of clusters, $K$, is dynamically determined, satisfying $1 \le K \le N$.

\subsection{Framework Overview}
\begin{figure*}[t!]
    \centering
    \includegraphics[width=\linewidth]{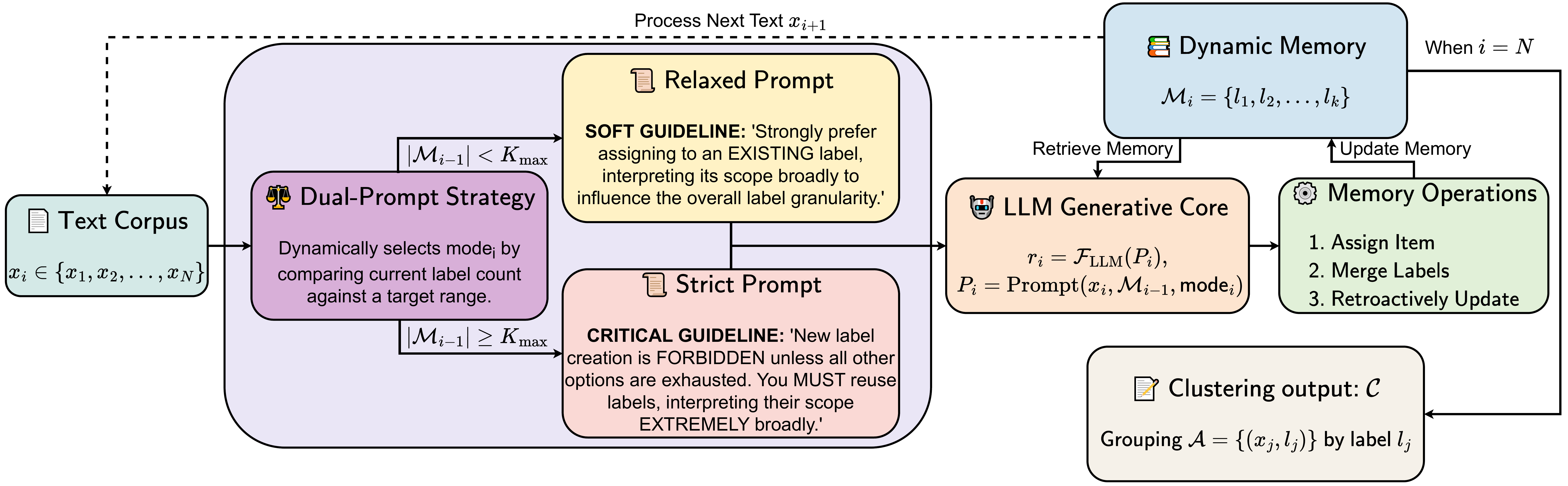}
    \caption{An overview of our proposed LLM-MemCluster framework. This figure illustrates the core iterative process, which is driven by a Dynamic Memory mechanism and the Dual-Prompt Strategy.}
    \label{fig:framework}
\end{figure*}

We propose \textbf{LLM-MemCluster}, a novel framework that leverages API calls to a large language model (LLM), eliminating the need for model fine-tuning or integration with traditional algorithms. As illustrated in Figure~\ref{fig:framework}, our framework is designed to directly address two principal challenges: the statelessness of LLMs and the inherent ambiguity in determining the number of clusters.

The architecture of LLM-MemCluster is centered on two synergistic innovations: a \textbf{Dynamic Memory} mechanism that endows the LLM with a functional state, and a \textbf{Dual-Prompt Strategy} for active control over clustering granularity. The framework processes each text instance from $\mathcal{D}$ sequentially. Throughout this process, it maintains a dynamic set of assignments $\mathcal{A} = \{(x_j, l_j)\}_{j=1}^N$, recording the label $l_j$ for each instance $x_j$. This set $\mathcal{A}$ is crucial for the iterative refinement and is used to produce the final partition $\mathcal{C}$.

\subsection{Stateful Clustering via Dynamic Memory}

The inherent statelessness of contemporary LLMs, confining their operational memory to a single context window, presents a significant challenge for iterative tasks. In clustering, this leads to inconsistent assignments and redundant clusters. Our Dynamic Memory mechanism addresses this by providing the LLM with a persistent working memory of the evolving cluster landscape. Our framework operates in a single pass, efficiently completing the clustering of $N$ instances in exactly $N$ steps, unlike iterative methods like K-Means.

The memory module, denoted as $\mathcal{M}_{\text{mem}}$, maintains a dynamically updated set of descriptive labels representing the discovered clusters (e.g., {``Arts'', ``Science''}). At each step $i$ (for $i=1, \dots, N$), the framework processes instance $x_i$. Let $\mathcal{M}_{i-1}$ be the memory state before this step. The core operation is to invoke the LLM, modeled as a function $\mathcal{F}_{\text{LLM}}$, with a prompt constructed from the current instance $x_i$, the memory state $\mathcal{M}_{i-1}$, and the active mode (detailed in Section~\ref{ssec:dual_prompt}). Conceptually, the LLM returns a structured tuple containing an assignment label $l_i$ and an optional merge suggestion $s_i$. A merge suggestion has the form $(\mathcal{L}_{\text{old}}, l_{\text{new}})$; for example, $s_i = (\{ \text{``ML''}, \text{``DL''} \}, \text{``AI''})$ proposes consolidating existing labels. Formally:
\begin{equation}
    \label{eq:core_call}
    (l_i, s_i) = \mathcal{F}_{\text{LLM}}(x_i, \mathcal{M}_{i-1}, \text{mode}_i).
\end{equation}

The framework uses the returned label $l_i$ and merge suggestion $s_i$ to update its state, cycling through three core operations:

\begin{itemize}
    \item \textbf{Reuse or Create.} For each text instance $x_i$, our framework constructs a prompt containing the current set of labels from $\mathcal{M}_{i-1}$. The LLM's primary directive is to either reuse an existing label for $x_i$ or create a new label if the text represents a fundamentally distinct topic, yielding the intermediate memory state $\mathcal{M}'_i$:
          \begin{equation}
              \label{eq:reuse_create}
              \mathcal{M}'_i =
              \begin{cases}
                  \mathcal{M}_{i-1} \cup \{l_i\} & \text{if } l_i \text{ is a new label} \\
                  \mathcal{M}_{i-1}              & \text{otherwise}
              \end{cases}
          \end{equation}

    \item \textbf{Merge and Refine.} A defining feature of our framework is its capacity to direct the LLM to propose a \texttt{MERGE\_SUGGESTION} at any step, enabling proactive consolidation of semantically similar or redundant labels. Crucially, this is an optional, concurrent action rather than a post-processing phase, allowing real-time optimization of the label space. The memory state $\mathcal{M}_i$ is updated based on the merge suggestion $s_i=(\mathcal{L}_{\text{old}}, l_{\text{new}})$:

          \begin{equation}
              \label{eq:merge_refine}
              \mathcal{M}_i =
              \begin{cases}
                  (\mathcal{M}'_i \setminus \mathcal{L}_{\text{old}}) \cup \{l_{\text{new}}\} & \text{on merge}  \\
                  \mathcal{M}'_i                                                              & \text{otherwise}
              \end{cases}
          \end{equation}

    \item \textbf{Retroactive Update.} Upon receiving a merge suggestion, the framework updates $\mathcal{M}_{\text{mem}}$ and retroactively re-maps historical assignments. This ensures global consistency (detailed in Appendix~\ref{app:algo_implementation}). Specifically, any assignment $(x_j, l_j) \in \mathcal{A}$ is updated to $(x_j, l'_j)$, where:
          \begin{equation}
              \label{eq:retroactive_update}
              l'_j =
              \begin{cases}
                  l_{\text{new}} & \text{if } l_j \in \mathcal{L}_{\text{old}} \\
                  l_j            & \text{otherwise}
              \end{cases}
          \end{equation}
\end{itemize}
This integrated cycle transforms the stateless LLM into a state-aware clustering agent, ensuring both local accuracy and global consistency. The process is driven by a structured prompt (see Appendix~\ref{app:prompt_details}), which instructs the LLM to return a primary assignment—either reusing or creating a label—and, optionally, a merge suggestion.

\subsection{Dual-Prompt Granularity Control}
\label{ssec:dual_prompt}

We introduce the Dual-Prompt Strategy to provide users with a means of actively guiding the final clustering granularity. This approach addresses the canonical challenge of steering the final number of clusters ($K$) to align with user-defined goals, and is implemented as a dedicated control layer that actively regulates the cluster count. By doing so, the strategy ensures the final partition conforms to user expectations or the data's intrinsic structure.

This strategy modulates the LLM's propensity for new label creation by dynamically switching between two prompting modes. The mechanism is guided by a user-defined target range for the cluster count, $[K_{\min}, K_{\max}]$. While the entire range is provided to the LLM as a contextual guideline for its decision-making, the programmatic switch between modes is triggered by the upper bound, $K_{\max}$. Specifically, the prompt mode for instance $x_i$ is determined by the current cluster count:
\begin{equation}
    \label{eq:mode_switch}
    \text{mode}_i =
    \begin{cases}
        \text{Strict}  & \text{if } |\mathcal{M}_{i-1}| \ge K_{\max} \\
        \text{Relaxed} & \text{otherwise}
    \end{cases}
\end{equation}
This strategy uses two distinct prompt templates:
\begin{enumerate}
    \item \textbf{The Strict Prompt:} Activated when the current number of clusters meets or exceeds the desired maximum, this mode incorporates prescriptive constraints into the prompt, significantly curtailing new label creation and compelling the LLM to prioritize \textbf{Reuse} and \textbf{Merge}. This raises the threshold for creating new clusters, hindering their formation.
    \item \textbf{The Relaxed Prompt:} As the default operational mode, this prompt is used when the cluster count is within the desired range. It grants the LLM greater latitude in label creation, allowing it to form new clusters for semantically distinct topics as needed, thereby facilitating cluster discovery.
\end{enumerate}
By adjusting prompt constraints based on the real-time cluster count, this strategy provides explicit control over the final clustering granularity, preventing uncontrolled label growth or premature consolidation. The complete strict and relaxed prompt templates are detailed in Appendix~\ref{app:prompt_details}. Finally, we provide the detailed algorithmic pseudocode and a computational complexity analysis in Appendix~\ref{app:algo_implementation}.

\section{Experiments}

In this section, we evaluate our proposed framework, \textbf{LLM-MemCluster}, through experiments addressing the following research questions:
\begin{itemize}
    \item \textbf{RQ1:} How does LLM-MemCluster perform against a variety of strong clustering baselines that employ different algorithms and state-of-the-art text representations?
    \item \textbf{RQ2:} How do the Dynamic Memory and Dual-Prompt Strategy components individually contribute to the overall effectiveness of our proposed LLM-MemCluster?
    \item \textbf{RQ3:} How robust is the LLM-MemCluster framework to variations in experimental conditions, specifically the dual-prompt transition threshold and the dataset execution order?
    \item \textbf{RQ4:} What is the generalization capability of the LLM-MemCluster framework when its foundational component is substituted with different large language models?
\end{itemize}

\subsection{Experimental Setup}

\subsubsection{Datasets}

We evaluate our method on six public benchmark datasets~\cite{zhang-etal-2023-clusterllm}, selected to cover a wide range of text clustering challenges. As detailed in Appendix~\ref{app:dataset_table}, these datasets span numerous domains and feature a broad range of cluster counts (K from 18 to 102), providing a robust testbed to assess the generalization of our method.

\subsubsection{Evaluation Metrics}
We evaluate performance using three standard metrics, where higher values indicate better performance and 1 denotes a perfect score:
\begin{itemize}
    \item \textbf{Accuracy (ACC):} Calculates the percentage of correctly assigned data points, based on the optimal one-to-one mapping between predicted clusters and ground-truth labels.
    \item \textbf{Normalized Mutual Information (NMI):} Measures the mutual information between predicted and true labels, normalized by their entropies. It quantifies the statistical information shared between the two assignments.
    \item \textbf{Adjusted Rand Index (ARI):} A chance-adjusted measure of similarity between two data clusterings. It is calculated based on the proportion of sample pairs that are correctly assigned to the same or different clusters.
\end{itemize}

\subsubsection{Baselines}
We assess effectiveness by benchmarking against baselines from three distinct paradigms:

\begin{itemize}
    \item \textbf{Traditional Method:} K-Means on TF-IDF vectors, a classic baseline relying on sparse, high-dimensional lexical features.
    \item \textbf{Embedding-based Methods:} We evaluate algorithms representing three key approaches: the centroid-based K-Means~\cite{lloyd1982least}, the density-based DBSCAN~\cite{deng2020dbscan}, and the graph-based Spectral Clustering~\cite{ng2001spectral}. We apply these methods to \texttt{instructor-large} embeddings~\cite{su-etal-2023-one} alongside the established BERTopic pipeline~\cite{grootendorst2022bertopic}.
    \item \textbf{LLM-based Method:} We compare against ClusterLLM~\cite{zhang-etal-2023-clusterllm}, which uses an LLM to generate pseudo-labels for training a smaller sentence encoder, enabling a highly scalable, multi-stage clustering approach. To ensure reproducibility, we set the temperature to 0 for all LLM-based experiments.
\end{itemize}

\subsection{Main Results (RQ1)}

\begin{table*}[t!]
    \centering
    \setlength{\tabcolsep}{2.5pt}
    \resizebox{\textwidth}{!}{%
        \begin{tabular}{l ccc ccc ccc ccc ccc ccc ccc}
            \toprule
            \multirow{2}{*}{Method} & \multicolumn{3}{c}{ArxivS2S} & \multicolumn{3}{c}{Massive-I} & \multicolumn{3}{c}{MTOP-I} & \multicolumn{3}{c}{Massive-D} & \multicolumn{3}{c}{FewNerd} & \multicolumn{3}{c}{FewRel} & \multicolumn{3}{c}{AVG}                                                                                                                                                                                                                                                                           \\
            \cmidrule(lr){2-4} \cmidrule(lr){5-7} \cmidrule(lr){8-10} \cmidrule(lr){11-13} \cmidrule(lr){14-16} \cmidrule(lr){17-19} \cmidrule(lr){20-22}
                                    & ACC                          & NMI                           & ARI                        & ACC                           & NMI                         & ARI                        & ACC                     & NMI              & ARI              & ACC              & NMI              & ARI              & ACC              & NMI              & ARI              & ACC              & NMI              & ARI              & ACC              & NMI              & ARI              \\
            \midrule
            K-Means-TF-IDF          & 12.1                         & 31.7                          & 1.2                        & 31.1                          & 49.8                        & 8.4                        & 31.7                    & 55.3             & 16.0             & 43.9             & 44.2             & 10.5             & 11.1             & 37.7             & 1.0              & 23.1             & 36.1             & 4.2              & 25.5             & 42.5             & 6.9              \\
            DBSCAN                  & 6.3                          & 17.6                          & 0.3                        & 20.4                          & 28.9                        & 1.0                        & 21.5                    & 26.6             & 2.2              & 25.9             & 30.7             & 6.4              & 27.0             & 0.5              & 0.1              & 10.5             & 19.5             & 0.4              & 18.6             & 20.6             & 1.7              \\
            Spectral                & \underline{25.1}             & 48.0                          & 9.2                        & \textbf{60.5}                 & 72.9                        & 38.7                       & \underline{39.2}        & 68.8             & 27.8             & 54.1             & 64.7             & 33.0             & 34.0             & 42.0             & 9.3              & 35.4             & 51.5             & 15.7             & 41.4             & 58.0             & 22.3             \\
            BERTopic                & 17.9                         & 39.2                          & 1.8                        & 52.5                          & 70.0                        & 32.5                       & 35.8                    & 64.1             & 15.9             & 52.6             & 56.2             & 29.3             & 34.9             & 40.3             & \underline{11.7} & 31.1             & 50.7             & 9.7              & 37.5             & 53.4             & 16.8             \\
            K-Means-Inst            & \underline{25.1}             & 49.3                          & 12.3                       & \underline{55.7}              & 72.6                        & 41.6                       & 34.5                    & 70.9             & 26.9             & \underline{54.9} & \underline{66.9} & \underline{42.7} & 28.2             & 43.3             & 6.1              & 34.8             & 53.1             & 22.5             & 38.9             & 59.3             & 25.4             \\
            ClusterLLM              & \underline{25.1}             & \underline{50.5}              & \underline{13.7}           & 55.5                          & \textbf{74.6}               & \underline{43.2}           & 36.0                    & \underline{73.4} & \underline{30.0} & 52.4             & 65.3             & 40.8             & \underline{37.3} & \underline{53.1} & 10.7             & \textbf{43.8}    & \underline{59.6} & \underline{30.4} & \underline{41.7} & \underline{62.8} & \underline{28.1} \\
            \textbf{Our Method}     & \textbf{28.4}                & \textbf{57.4}                 & \textbf{16.3}              & 54.8                          & \underline{73.5}            & \textbf{47.9}              & \textbf{64.0}           & \textbf{77.5}    & \textbf{68.9}    & \textbf{57.6}    & \textbf{67.7}    & \textbf{53.8}    & \textbf{59.3}    & \textbf{63.3}    & \textbf{53.1}    & \underline{43.2} & \textbf{63.6}    & \textbf{32.7}    & \textbf{51.2}    & \textbf{67.2}    & \textbf{45.4}    \\
            \bottomrule
        \end{tabular}%
    }
    \caption{Comparison of LLM-MemCluster with baselines across six datasets using ACC, NMI, and ARI scores (\%). The best and second-best results are highlighted in \textbf{bold} and \underline{underlined}, respectively. Baselines utilize \texttt{instructor-large} embeddings (except K-Means-TF-IDF), while our method employs in-context learning and ClusterLLM uses it to provide guidance (both utilizing the GPT-4.1 mini model).}
    \label{tab:main_results}
\end{table*}

As shown in Table~\ref{tab:main_results}, our framework, LLM-MemCluster, establishes a new state-of-the-art in unsupervised text clustering. On average, LLM-MemCluster surpasses the strongest baseline, ClusterLLM, by absolute margins of 9.5\% in ACC, 4.4\% in NMI, and 17.3\% in ARI. The framework's advantages are particularly evident on high-cardinality datasets where conventional methods tend to falter. For instance, on MTOP-I (K=102), it achieves an ARI of 68.9---a 38.9-point improvement over ClusterLLM. A similar 42.4-point gain in ARI on FewNerd (K=58) further demonstrates its effectiveness for semantically complex tasks.

These results offer a crucial insight: superior clustering performance is not merely a function of powerful text representations, but rather a result of an architectural design that effectively leverages these representations. This architectural dependence highlights the fundamental limitations of baseline methods. Embedding-based approaches, such as Spectral Clustering, rely on static vectors that, despite their quality, lack contextual adaptability. Other LLM-based methods like ClusterLLM treat the LLM as an external guide for knowledge distillation, rather than as a dynamic agent within the clustering process. This claim is further substantiated by our comprehensive generalization experiments in RQ4 (Section~\ref{ssec:gentollms}).

In contrast, the success of LLM-MemCluster is rooted in its novel architecture, which engages the LLM as a direct and active agent within a stateful, iterative process. The dynamic memory mechanism enables the framework to build a coherent, evolving understanding of the cluster space. This, in turn, allows the LLM to make adaptive, context-aware decisions at each step. We argue this direct and dynamic orchestration of the LLM's decision-making is the key innovation, allowing our method to navigate nuanced semantic relationships for more robust and accurate clustering.

\subsection{Ablation Study (RQ2)}\label{ssec:ablation}

\begin{figure*}[t!]
    \centering
    \begin{subfigure}[b]{0.36\textwidth}
        \centering
        \includegraphics[width=\textwidth]{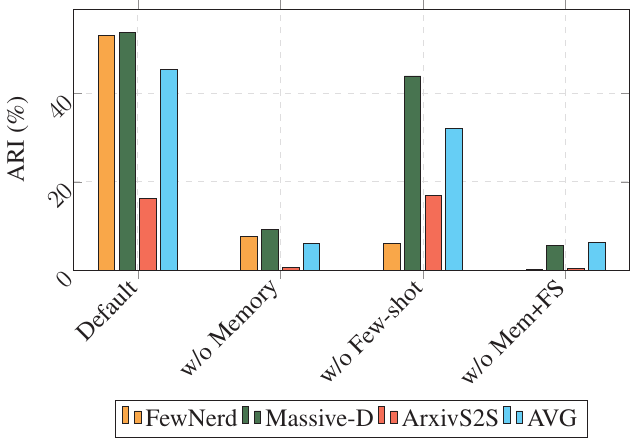}
        \caption{Memory \& Few-shot Ablation}
        \label{fig:ablation_memory}
    \end{subfigure}
    \hspace{0.05\textwidth}
    \begin{subfigure}[b]{0.36\textwidth}
        \centering
        \includegraphics[width=\textwidth]{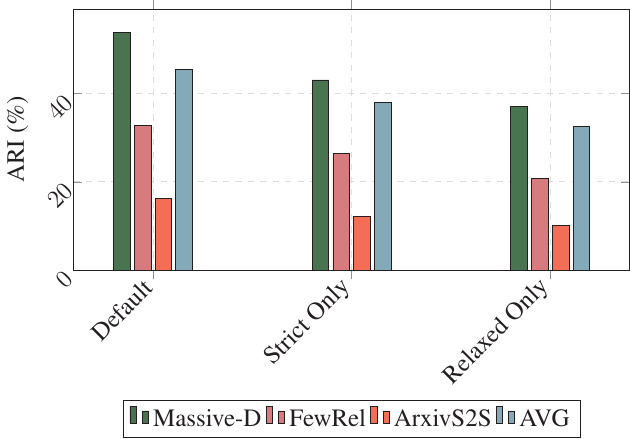}
        \caption{Prompt Strategy Ablation}
        \label{fig:ablation_prompt}
    \end{subfigure}

    \begin{subfigure}[b]{0.25\textwidth}
        \centering
        \includegraphics[width=\textwidth]{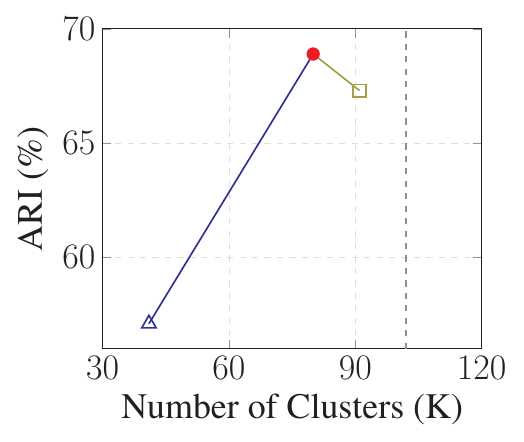}
        \caption{MTOP-I}
        \label{fig:mtop_i}
    \end{subfigure}
    \hfill
    \begin{subfigure}[b]{0.25\textwidth}
        \centering
        \includegraphics[width=\textwidth]{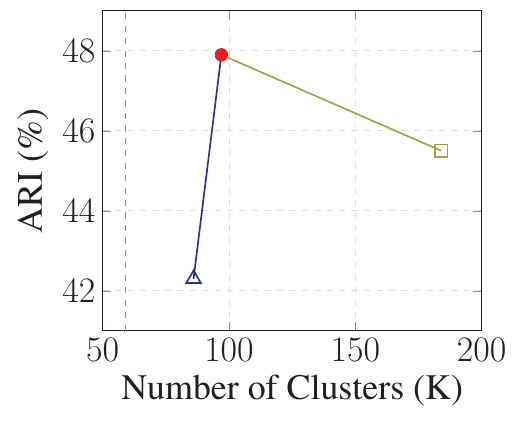}
        \caption{Massive-I}
        \label{fig:massive_i}
    \end{subfigure}
    \hfill
    \begin{subfigure}[b]{0.24\textwidth}
        \centering
        \includegraphics[width=\textwidth]{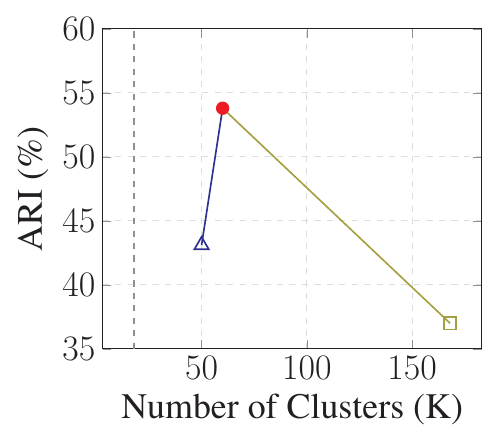}
        \caption{Massive-D}
        \label{fig:massive_d}
    \end{subfigure}
    \hfill
    \begin{subfigure}[b]{0.24\textwidth}
        \centering
        \includegraphics[width=\textwidth]{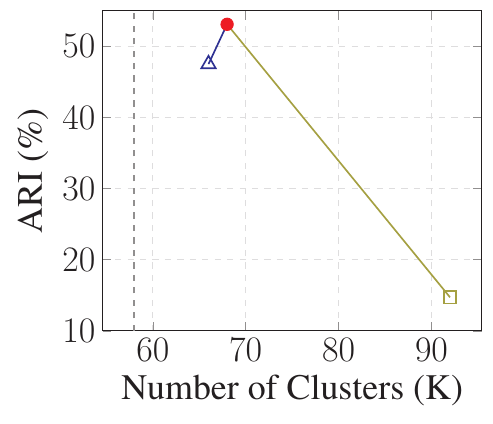}
        \caption{FewNerd}
        \label{fig:fewnerd}
    \end{subfigure}
    \centerline{\includegraphics[width=0.6\textwidth]{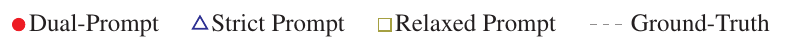}}
    \caption{Comprehensive ablation study and adaptive clustering strategy comparison.}
    \label{fig:combined_ablation}
\end{figure*}

We conduct a comprehensive study to validate the contributions of our framework's modules. The primary findings are summarized in Figure~\ref{fig:combined_ablation} and analyzed in detail in the subsequent subsections. Full numerical breakdowns for all experimental variants are provided in Appendix~\ref{app:ablation_table}.

\textbf{Memory and Grounding are Indispensable.} Figure~\ref{fig:ablation_memory} starkly highlights the critical roles of memory and in-context examples. Deactivating the Dynamic Memory (\textit{w/o Memory}) causes a catastrophic performance degradation across all datasets, validating that an external memory is essential for overcoming LLM statelessness. The importance of grounding the model with few-shot examples is also evident, though its impact varies. Removing them (\textit{w/o Few-shot}) generally leads to a significant ARI drop, a trend mirrored in Massive-D (from 53.8 to 44.0). However, the effect is exceptionally pronounced on FewNerd, where the ARI collapses from 53.1 to a mere 6.1. The stark performance drop on FewNerd underscores that for semantically complex domains, few-shot grounding is a prerequisite for robust performance.

\textbf{The Dual-Prompt Strategy is Highly Effective.} As shown in Figure~\ref{fig:ablation_prompt}, the superiority of our dual-prompt approach is evident, as variants relying on a single prompt type consistently underperform the full model. This pattern is not only clear on average—where the full model achieves a 45.4 ARI, compared to 38.1 for the strict-only and 32.5 for the relaxed-only variants—but is also robustly replicated across individual datasets. For instance, on Massive-D the full model's 53.8 ARI significantly exceeds the alternatives (43.1 and 37.0); a similar trend is observed on FewRel (32.7 vs. 26.5 and 20.8). This consistent underperformance validates our core design principle: a dynamic transition from an exploratory to a consolidative phase is the most effective strategy for reliably achieving optimal clustering granularity.

To further analyze how optimal granularity is achieved, Figures~\ref{fig:mtop_i} to~\ref{fig:fewnerd} illustrate the framework's adaptive behavior on representative datasets. On Massive-I, the framework achieves a higher ARI score via \textbf{semantic splitting}, producing more clusters than the ground-truth by identifying fine-grained sub-topics. Conversely, on the high-cardinality MTOP-I dataset, it performs \textbf{semantic consolidation}, merging overly similar categories to produce fewer clusters. Crucially, in both scenarios, the Dual-Prompt strategy yields the solution with the highest ARI score. This demonstrates that the framework does not rigidly pursue a specific K but rather optimizes for semantic coherence, adaptively deciding whether to split or merge, a determination strictly contingent on the intrinsic semantic properties of each dataset.

\subsection{Robustness Analysis (RQ3)}\label{ssec:hyperparam}
\begin{figure*}[t!]
    \centering
    \begin{subfigure}[b]{0.28\textwidth}
        \centering
        \includegraphics[width=\linewidth]{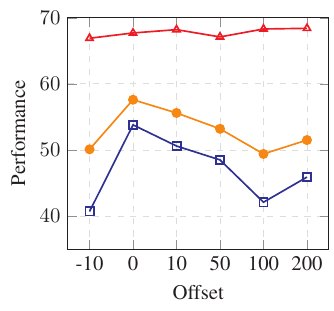}
        \caption{Massive-D}
        \label{fig:hyper_massived}
    \end{subfigure}
    \hfill
    \begin{subfigure}[b]{0.28\textwidth}
        \centering
        \includegraphics[width=\linewidth]{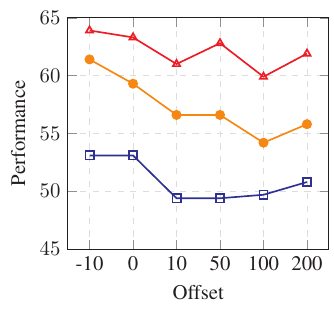}
        \caption{FewNerd}
        \label{fig:hyper_fewnerd}
    \end{subfigure}
    \hfill
    \begin{subfigure}[b]{0.28\textwidth}
        \centering
        \includegraphics[width=\linewidth]{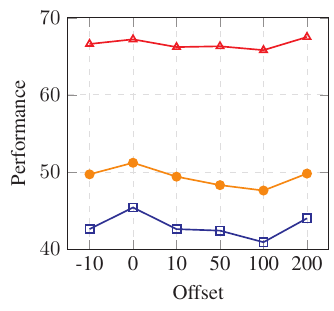}
        \caption{Average}
        \label{fig:hyper_avg}
    \end{subfigure}
    \centerline{\includegraphics[width=0.3\textwidth]{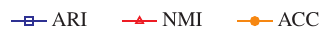}}
    \caption{Hyperparameter sensitivity analysis of the prompt transition threshold, demonstrating robust and near-optimal performance across a wide range of values for representative datasets and on average.}
    \label{fig:hyperparam_robustness}
\end{figure*}

To address RQ3, we evaluate the robustness of LLM-MemCluster against variations in experimental conditions, specifically focusing on the hyperparameter sensitivity and the stability under randomized dataset execution orders.

\paragraph{Hyperparameter Sensitivity}
We first analyze the sensitivity to the core hyperparameter: the transition threshold for the Dual-Prompt Strategy. This threshold determines when the model switches from the initial, exploratory relaxed prompt to the subsequent, consolidative strict prompt. We operationalize this threshold as an offset applied to the upper bound of the target range ($K_{max}$). A positive offset extends the exploratory phase, while a negative offset accelerates the consolidation process.

We evaluated a broad spectrum of offsets: -10, 0, +10, +50, +100, and +200. The results are visualized in Figure~\ref{fig:hyperparam_robustness}, which plots performance against different threshold offsets, with detailed results in Appendix~\ref{app:hyperparam_sensitivity}. While the average performance across all datasets (Figure~\ref{fig:hyper_avg}) shows relatively flat curves, this stability is even more evident on individual datasets. For instance, on FewNerd (Figure~\ref{fig:hyper_fewnerd}), the ARI score is exceptionally stable, fluctuating only minimally between a peak of 53.1 (default offset) and a low of 49.4. Even on Massive-D (Figure~\ref{fig:hyper_massived}), which exhibits more variance, performance peaks at an offset of 0 (53.8 ARI) and remains competitive across a wide range.

Notably, the framework performs well even at the extremes. An extended exploratory phase (offset +200) yields a strong ARI of 44.0 and the highest average NMI of 67.5. Conversely, an accelerated transition (offset -10) also maintains a robust 42.6 ARI. This resilience at the boundaries, mirrored across representative datasets, highlights the inherent robustness and self-correcting capacity of the dual-prompt mechanism, which adapts effectively to minor variations in the consolidation timing and control process.

\paragraph{Seed Robustness}
Beyond hyperparameter settings, we also examine the stability of our method under different dataset execution orders. This is a critical factor for stream clustering algorithms, where the processing sequence can inherently influence the resulting cluster integrity. We conducted experiments using 5 random seeds to simulate permuted input streams and report the mean and standard deviation in Appendix Table~\ref{tab:seed_robustness_gemini25_flash_lite}. The results demonstrate that LLM-MemCluster maintains consistent performance regardless of the input order. The standard deviation of the ARI scores remains low across all datasets, specifically falling within a tight 5\% margin. For example, on Massive-I and MTOP-I, the standard deviations are merely 2.85\% and 3.75\%, respectively. This low variance statistically confirms that our dynamic memory and adaptive dual-prompt mechanisms work in concert to resiliently capture global semantic structures independent of the local input sequence.

Consequently, our framework provides a practical advantage by delivering near-optimal, reproducible results across diverse datasets without requiring dataset-specific tuning or seed selection.

\subsection{Generalization to Different LLMs (RQ4)}\label{ssec:gentollms}

\begin{table*}[t!]
    \centering
    \setlength{\tabcolsep}{2.5pt}
    \resizebox{\textwidth}{!}{%
        \begin{tabular}{l ccc ccc ccc ccc ccc ccc ccc}
            \toprule
            \multirow{2}{*}{Base LLM} & \multicolumn{3}{c}{ArxivS2S} & \multicolumn{3}{c}{Massive-I} & \multicolumn{3}{c}{MTOP-I} & \multicolumn{3}{c}{Massive-D} & \multicolumn{3}{c}{FewNerd} & \multicolumn{3}{c}{FewRel} & \multicolumn{3}{c}{AVG}                                                                                                   \\
            \cmidrule(lr){2-4} \cmidrule(lr){5-7} \cmidrule(lr){8-10} \cmidrule(lr){11-13} \cmidrule(lr){14-16} \cmidrule(lr){17-19} \cmidrule(lr){20-22}
                                      & ACC                          & NMI                           & ARI                        & ACC                           & NMI                         & ARI                        & ACC                     & NMI  & ARI  & ACC  & NMI  & ARI  & ACC  & NMI  & ARI  & ACC  & NMI  & ARI  & ACC  & NMI  & ARI  \\
            \midrule
            \textbf{GPT-4.1-M}        & 28.4                         & 57.4                          & 16.3                       & 54.8                          & 73.5                        & 47.9                       & 64.0                    & 77.5 & 68.9 & 57.6 & 67.7 & 53.8 & 59.3 & 63.3 & 53.1 & 43.2 & 63.6 & 32.7 & 51.2 & 67.2 & 45.4 \\
            \midrule
            GPT-3.5-T                 & 35.9                         & 59.5                          & 19.8                       & 43.5                          & 70.0                        & 38.2                       & 52.6                    & 75.0 & 54.0 & 55.0 & 65.4 & 45.1 & 42.0 & 62.9 & 24.8 & 25.2 & 49.9 & 14.6 & 42.4 & 63.8 & 32.8 \\
            GPT-4.1                   & 29.8                         & 60.4                          & 18.6                       & 64.0                          & 77.3                        & 54.6                       & 67.8                    & 80.9 & 69.3 & 52.8 & 69.5 & 47.3 & 59.1 & 73.4 & 51.1 & 48.4 & 69.4 & 30.8 & 53.6 & 71.8 & 45.3 \\
            Gemini-2.0-F              & 33.8                         & 60.2                          & 21.3                       & 29.1                          & 51.7                        & 11.1                       & 63.2                    & 74.7 & 63.4 & 50.0 & 63.2 & 35.9 & 46.3 & 57.4 & 52.3 & 37.1 & 59.1 & 23.9 & 43.3 & 61.1 & 34.6 \\
            DeepSeek-V3               & 23.9                         & 54.1                          & 14.9                       & 48.1                          & 69.1                        & 39.3                       & 60.9                    & 73.7 & 63.2 & 53.7 & 59.7 & 39.7 & 53.6 & 62.0 & 62.3 & 20.9 & 46.5 & 12.7 & 43.5 & 60.9 & 38.7 \\
            Gemini-2.5-F              & 34.8                         & 68.4                          & 24.9                       & 50.6                          & 77.3                        & 41.6                       & 60.3                    & 80.4 & 60.5 & 54.5 & 67.4 & 43.8 & 65.7 & 76.4 & 59.4 & 48.3 & 70.4 & 36.9 & 52.4 & 73.4 & 44.5 \\
            \bottomrule
        \end{tabular}%
    }
    \caption{Generalization of the framework across large language models in ACC, NMI, and ARI (\%). For brevity, we abbreviate model names: GPT-4.1-M (GPT-4.1-mini), GPT-3.5-T (GPT-3.5-turbo), Gemini-2.0-F (Gemini-2.0-flash), DeepSeek-V3 (DeepSeek-V3-0324), and Gemini-2.5-F (Gemini-2.5-flash-preview-05-20).}
    \label{tab:llm_generalization}
\end{table*}

In addressing RQ4, we assess our framework's generalization by evaluating it across a range of Large Language Models, including GPT-4.1-mini (default), GPT-3.5-turbo, GPT-4.1, Gemini-2.0-flash, Gemini-2.5-flash-preview-05-20, and DeepSeek-V3-0324, thereby confirming its portability.

The results, presented in Table~\ref{tab:llm_generalization}, highlight the framework's strong portability and robustness. Performance remains exceptionally strong when other high-capability models are used. For instance, substituting our default GPT-4.1-mini (45.4 ARI) with the more powerful GPT-4.1 yields a nearly identical ARI of 45.3. Similarly, competitive performance is observed with Gemini-2.5-flash-preview-05-20 (44.5 ARI). The strong results from other models, including DeepSeek-V3-0324 (38.7 ARI), confirm our design's successful generalization across a diverse set of LLM backbones.

The advantages of our design are most apparent when paired with less capable models. For instance, our framework achieves a 34.6 ARI using Gemini-2.0-flash, significantly surpassing the 28.1 ARI of ClusterLLM, which uses the more capable GPT-4.1-mini (Tables~\ref{tab:llm_generalization} and~\ref{tab:main_results}). This comparison strongly indicates that our innovative design approach, rather than the underlying model's intrinsic capability, primarily accounts for the gains.

Beyond performance, we provide a detailed analysis of the monetary cost in Appendix~\ref{app:token_cost}. A key efficiency advantage of our design is that token consumption is predominantly driven by the input context, whereas the output completion tokens remain minimal. Given that completion tokens often incur higher rates, this characteristic ensures that monetary costs remain manageable and predictable, even when scaling to premium models.

\section{Related Work}

\subsection{LLM-Augmented Clustering}
A prominent approach employs a Large Language Model (LLM) as a high-level ``oracle'' to augment or refine clustering pipelines that rely on external models. These methods distill the LLM's semantic judgment to address specific, challenging parts of the clustering process. For instance, Cequel~\cite{10.1145/3746252.3761074} generates pairwise constraints~\cite{basu2004active} to guide a downstream clustering algorithm. Other work focuses on refinement, where LLMEdgeRefine~\cite{feng-etal-2024-llmedgerefine} re-assigns ambiguous ``edge points'' on the boundaries of initial clusters to enhance their integrity. A third approach, ClusterLLM~\cite{zhang-etal-2023-clusterllm}, leverages an LLM to generate supervisory signals from confusing document triplets to effectively fine-tune a smaller, more efficient sentence encoder. While pragmatic, these ``LLM-as-oracle'' frameworks are hybrid solutions and do not constitute an end-to-end generative clustering process.

\subsection{End-to-End Generative Clustering}
A more recent paradigm shift leverages LLMs as standalone clustering agents, bypassing traditional numerical algorithms. A representative approach within this paradigm reframes clustering as a classification problem. The T-CLC framework~\cite{10.1145/3767695.3769519}, for example, operates in two distinct stages where it first prompts an LLM to generate candidate labels. However, this approach relies on a subset of ground-truth labels to seed the generation process, effectively shifting the task paradigm towards semi-supervised learning. In contrast, our work addresses the more challenging, fully unsupervised setting where no prior knowledge of the label space is available. Other related works have focused on improving the prompting process. ZeroDL~\cite{jo-etal-2025-zerodl}, for instance, first performs an open-ended inference step to learn the dataset's underlying distribution and then incorporates this meta-knowledge into a more data-aware prompt. However, this approach still treats clustering as a static inference task rather than a dynamic process with a continuously evolving state.

Our work, LLM-MemCluster, builds upon a generative paradigm but introduces a novel framework designed for efficient, iterative, single-pass clustering. In contrast to the multi-stage or static-inference approaches, it employs a \textbf{Dynamic Memory} to create a stateful process that continuously refines the cluster space~\cite{xu2025mem,11132479}. Furthermore, our \textbf{Dual-Prompt Strategy} provides an explicit mechanism for active granularity control throughout the process, addressing the challenge of dynamically determining the cluster count~\cite{petnehazi2025hercules}.
\section{Conclusion}

We introduce LLM-MemCluster, an end-to-end text clustering framework using a Dynamic Memory and Dual-Prompt Strategy to operate as a stateful, iterative agent, addressing the core challenges of LLM statelessness and cluster granularity. Its robust architecture achieves state-of-the-art performance, establishing an LLM-native paradigm that advances beyond hybrid approaches to effectively unlock LLM potential in unsupervised tasks.

\section*{Limitations}
Our framework entails specific trade-offs. First, it relies on the instruction-following capabilities of the LLM; while our approach is effective on advanced models, performance may vary on smaller architectures unable to strictly adhere to complex dual-prompt constraints. Second, as is inherent to its single-pass streaming nature, it optimizes based on the evolving memory state rather than performing multi-pass global refinement over the entire dataset. Finally, although the user-defined target range offers flexible granularity control, it necessitates minimal domain knowledge regarding the expected cluster distribution.

\bibliography{anthology,custom}
\appendix

\section{Unified Prompt Template}
\label{app:prompt_details}

This section details the unified prompt template at the core of our framework. As shown in Figure~\ref{fig:prompt_template_main}, the template integrates our \textbf{Dynamic Memory} by injecting the current \texttt{Known labels}, and our \textbf{Dual-Prompt Strategy} via placeholders for dynamic instructions. The specific content for the \texttt{[SYSTEM\_GUIDELINE]} and \texttt{[USER\_CONSTRAINT]} placeholders is provided in Figures~\ref{fig:prompt_template_system_content} and~\ref{fig:prompt_template_user_content}.

\begin{figure*}[t!]
  \centering
  {\lstset{basicstyle=\ttfamily\footnotesize}
    \begin{lstlisting}
--- SYSTEM PROMPT ---
You are an expert text analysis and clustering specialist. 
Your primary goal is to determine the underlying theme, topic, 
or relation type for each text input and assign it to an 
appropriate category.

CORE PRINCIPLES:
- HIGHEST PRIORITY: Reuse existing labels whenever reasonably 
  possible to ensure consistency.
- NEW LABELS: Create ONLY AS A LAST RESORT when an input is
  FUNDAMENTALLY NEW.
- MERGE: Suggest merging similar labels to improve conciseness.

[SYSTEM_GUIDELINE]
    \end{lstlisting}
  }
  \caption{The unified prompt template (system prompt).}
  \label{fig:prompt_template_main}
\end{figure*}

\begin{figure*}[t!]
  \centering
  {\lstset{basicstyle=\ttfamily\footnotesize}
    \begin{lstlisting}
--- USER PROMPT ---
Known labels: ["label_1", "label_2", ...]

Examples:
Input: "Example text 1" -> Output: ASSIGNED_LABEL: "label_A"
Input: "Example text 2" -> Output: NEW_LABEL: "label_B"

Input to process: "text_to_cluster"

Instructions:
Your response must contain exactly one of the following primary lines:
- ASSIGNED_LABEL: <label_name>
- NEW_LABEL: <new_label_name> [USER_CONSTRAINT]

Optionally, you can also include the following line for consolidation:
- MERGE_SUGGESTION: MERGE: ["old_label"] INTO: ["new_label"]

RESPONSE FORMATTING:
- Exactly ONE 'ASSIGNED_LABEL:' OR 'NEW_LABEL:' line.
- Optionally, ONE 'MERGE_SUGGESTION:' line.
    \end{lstlisting}
  }
  \caption{The unified prompt template (user prompt, with \texttt{Known labels} from dynamic memory).}
  \label{fig:prompt_template_user}
\end{figure*}

\begin{figure*}[t!]
  \centering
  {\lstset{basicstyle=\ttfamily\footnotesize}
    \begin{lstlisting}
[SYSTEM_GUIDELINE] Content
--------------------------
# Relaxed Mode (Default)
SOFT GUIDELINE: As an additional consideration, try to manage the overall list of known labels such that the total number of unique labels ideally stays {range_desc}. This is a soft guideline to influence label granularity; your primary decision-making process (prioritize reuse, create new only if essential, suggest useful merges) remains paramount.

# Strict Mode
CRITICAL GUIDELINE: The total number of unique labels MUST be managed towards {range_desc}. If approaching/exceeding the upper limit, new label creation is SEVERELY RESTRICTED. You MUST aggressively reuse existing labels (interpret their scope VERY broadly) and proactively seek merge opportunities.
    \end{lstlisting}
  }
  \caption{Content for placeholder [SYSTEM\_GUIDELINE]. Injected into Figure~\ref{fig:prompt_template_main} based on the mode.}
  \label{fig:prompt_template_system_content}
\end{figure*}

\begin{figure*}[t!]
  \centering
  {\lstset{basicstyle=\ttfamily\footnotesize}
    \begin{lstlisting}
[USER_CONSTRAINT] Content
-------------------------
# Relaxed Mode
CONSIDERATION: If current known labels approach or exceed {target_max_clusters}, please be very cautious about creating NEW_LABEL. Strongly prefer assigning to an EXISTING label (interpret its scope broadly) or identifying a MERGE.

# Strict Mode
CRITICAL CHECK: If current known labels approach or exceed {target_max_clusters}, creating a NEW_LABEL is FORBIDDEN unless all other options are exhausted. You MUST first attempt to assign to an EXISTING label (interpret its scope EXTREMELY broadly) or identify a MERGE. Only if the input is unequivocally unique and NO existing label can accommodate it even with the broadest interpretation, and NO merge is possible, then, as a final resort, create a NEW_LABEL.
    \end{lstlisting}
  }
  \caption{Content for placeholder [USER\_CONSTRAINT]. Injected into Figure~\ref{fig:prompt_template_main} based on the mode.}
  \label{fig:prompt_template_user_content}
\end{figure*}

\subsection{Dynamic Placeholder Content}
\label{app:dynamic_content}
Our framework modulates the prompt's behavior by programmatically switching between two operational modes. Specifically, we dynamically populate two placeholders: \texttt{[SYSTEM\_GUIDELINE]} (system-level guidance) and \texttt{[USER\_CONSTRAINT]} (user-specific constraints), as defined in Figure~\ref{fig:prompt_template_main}. The actual content injected into these placeholders consists of the detailed instructions and constraints shown in Figures~\ref{fig:prompt_template_system_content} and~\ref{fig:prompt_template_user_content}. The transition between operational modes is governed by the number of discovered clusters relative to a user-defined upper bound, $K_{\max}$. The system operates in its \textbf{Relaxed Mode}, employing soft advisory language, as long as the cluster count remains below this threshold. Once $K_{\max}$ is reached or exceeded, the system transitions to \textbf{Strict Mode}, using restrictive language to enforce the cluster cardinality.

\section{Dataset Overview}
\label{app:dataset_table}

Table \ref{tab:datasets} lists the datasets used in our experiments, detailing each one's primary task or domain, the total number of samples, and the number of ground-truth clusters, denoted by $K$. We follow the experimental setup of \citet{zhang-etal-2023-clusterllm} and use their processed versions of these datasets. The original sources are as follows: ArxivS2S is adapted from the MTEB benchmark~\cite{muennighoff-etal-2023-mteb}, consisting of scientific abstracts; Massive-I and Massive-D are subsets of the MASSIVE dataset~\cite{fitzgerald-etal-2023-massive}, focusing on intent detection and domain classification, respectively; MTOP-I is derived from the MTOP benchmark~\cite{li-etal-2021-mtop}, specifically focusing on intent classification; FewNerd is a large-scale, fine-grained named entity recognition dataset~\cite{ding-etal-2021-nerd}; and FewRel is a benchmark dataset for few-shot relation classification~\cite{gao-etal-2019-fewrel}.

\begin{table}[t!]
  \centering
  \setlength{\tabcolsep}{2.5pt}
  \resizebox{\columnwidth}{!}{
    \begin{tabular}{llcc}
      \toprule
      Dataset   & Primary Task/Domain      & \# Samples & K   \\
      \midrule
      ArxivS2S  & Scientific Abstracts     & 3,674      & 93  \\
      Massive-I & Intent Detection         & 2,974      & 59  \\
      MTOP-I    & Intent Detection         & 4,386      & 102 \\
      Massive-D & Conversational Domain    & 2,974      & 18  \\
      FewNerd   & Named Entity Recognition & 3,789      & 58  \\
      FewRel    & Relation Extraction      & 4,480      & 64  \\
      \bottomrule
    \end{tabular}
  }
  \caption{Statistics of the datasets used in our experiments. K denotes the number of ground-truth clusters.}
  \label{tab:datasets}
\end{table}

\section{Detailed Ablation Study}
\label{app:ablation_table}

\begin{table*}[t!]
  \centering
  \setlength{\tabcolsep}{2.5pt}
  \resizebox{\textwidth}{!}{%
    \begin{tabular}{l ccc ccc ccc ccc ccc ccc ccc}
      \toprule
      \multirow{2}{*}{Method Variant} & \multicolumn{3}{c}{ArxivS2S} & \multicolumn{3}{c}{Massive-I} & \multicolumn{3}{c}{MTOP-I} & \multicolumn{3}{c}{Massive-D} & \multicolumn{3}{c}{FewNerd} & \multicolumn{3}{c}{FewRel} & \multicolumn{3}{c}{AVG}                                                                                                                                                                                                                                 \\
      \cmidrule(lr){2-4} \cmidrule(lr){5-7} \cmidrule(lr){8-10} \cmidrule(lr){11-13} \cmidrule(lr){14-16} \cmidrule(lr){17-19} \cmidrule(lr){20-22}
                                      & ACC                          & NMI                           & ARI                        & ACC                           & NMI                         & ARI                        & ACC                     & NMI           & ARI           & ACC           & NMI           & ARI           & ACC           & NMI           & ARI           & ACC           & NMI           & ARI           & ACC           & NMI           & ARI           \\
      \midrule
      \textbf{Default}                & \textbf{28.4}                & \textbf{57.4}                 & \textbf{16.3}              & \textbf{54.8}                 & \textbf{73.5}               & \textbf{47.9}              & \textbf{64.0}           & \textbf{77.5} & \textbf{68.9} & \textbf{57.6} & \textbf{67.7} & \textbf{53.8} & \textbf{59.3} & \textbf{63.3} & \textbf{53.1} & \textbf{43.2} & \textbf{63.6} & \textbf{32.7} & \textbf{51.2} & \textbf{67.2} & \textbf{45.4} \\
      \midrule
      w/o Memory                      & 4.7                          & 71.1                          & 0.5                        & 8.5                           & 65.2                        & 1.8                        & 11.2                    & 61.7          & 2.8           & 17.0          & 57.7          & 9.2           & 15.1          & 61.4          & 7.7           & 20.9          & 69.8          & 14.8          & 12.9          & 64.5          & 6.1           \\
      w/o Few-shot                    & 28.1                         & 58.5                          & 17.0                       & 50.2                          & 70.9                        & 43.5                       & 61.4                    & 74.3          & 60.5          & 55.5          & 65.7          & 44.0          & 27.3          & 48.7          & 6.1           & 33.5          & 55.3          & 21.2          & 42.7          & 62.2          & 32.1          \\
      w/o M+FS                        & 4.2                          & 71.0                          & 0.4                        & 18.5                          & 67.4                        & 11.9                       & 23.1                    & 65.2          & 18.5          & 11.7          & 55.4          & 5.5           & 4.3           & 57.2          & 0.2           & 5.3           & 66.2          & 1.2           & 11.2          & 63.7          & 6.3           \\
      Strict Prompt                   & 18.6                         & 49.0                          & 12.1                       & 52.6                          & 69.7                        & 42.3                       & 56.8                    & 74.8          & 57.1          & 50.2          & 67.6          & 43.1          & 57.8          & 62.6          & 47.5          & 38.1          & 64.1          & 26.5          & 45.7          & 64.6          & 38.1          \\
      Relaxed Prompt                  & 17.5                         & 60.6                          & 10.0                       & 54.7                          & 74.3                        & 45.5                       & 65.4                    & 78.3          & 67.3          & 41.7          & 65.4          & 37.0          & 39.4          & 52.1          & 14.7          & 33.7          & 56.6          & 20.8          & 42.1          & 64.5          & 32.5          \\
      \bottomrule
    \end{tabular}%
  }
  \caption{Ablation study of LLM-MemCluster supporting the analysis for RQ2. We report ACC, NMI, and ARI (\%), highlighting the importance of each component by comparing performance to the Default setting.}
  \label{tab:ablation_appendix}
\end{table*}

\begin{table}[t!]
  \centering
  \setlength{\tabcolsep}{2.5pt}
  \resizebox{\columnwidth}{!}{
    \begin{tabular}{l cccccc}
      \toprule
      Method Variant & ArxivS2S    & Massive-I   & MTOP-I       & Massive-D   & FewNerd     & FewRel      \\
      \midrule
      Ground-Truth   & \textbf{93} & \textbf{59} & \textbf{102} & \textbf{18} & \textbf{58} & \textbf{64} \\
      \midrule
      Dual-Prompt    & 159         & 97          & 80           & 60          & 68          & 122         \\
      Strict Prompt  & 70          & 86          & 41           & 50          & 66          & 89          \\
      Relaxed Prompt & 1208        & 184         & 91           & 168         & 92          & 141         \\
      \bottomrule
    \end{tabular}
  }
  \caption{Comparison of the number of clusters (K) produced by different model variants.}
  \label{tab:cluster_count_ablation}
\end{table}

\subsection{Analysis of Component Effectiveness}

Tables~\ref{tab:ablation_appendix} and~\ref{tab:cluster_count_ablation} provide the detailed quantitative analysis comprehensively supplementing the ablation study discussion in Section~\ref{ssec:ablation}.

\paragraph{The Roles of Memory and Grounding}
Our results in Table~\ref{tab:ablation_appendix} clearly validate that both Dynamic Memory and few-shot grounding are critical for the framework's overall success.
\begin{itemize}
  \item \textbf{Dynamic Memory (\texttt{w/o Memory}):} Deactivating the memory module leads to a near-total collapse in performance across all six datasets. The average ARI consequently plummets from \textbf{45.4\%} to a mere \textbf{6.1\%}. This confirms that an external, stateful memory is essential to overcome the inherent statelessness of LLMs for iterative tasks like clustering.
  \item \textbf{Few-shot Grounding (\texttt{w/o Few-shot}):} Removing the few-shot examples also causes a significant performance degradation, with the average ARI dropping from \textbf{45.4\%} to \textbf{32.1\%}. The effect is particularly dramatic on semantically nuanced datasets like FewNerd, where the ARI score collapses from \textbf{53.1\%} to just \textbf{6.1\%}. This highlights that for complex domains, providing in-context examples is crucial for effectively guiding the model to produce accurate and consistent outputs.
\end{itemize}

\paragraph{Effectiveness of the Dual-Prompt Strategy}

By conducting a cross-referenced analysis of the performance metrics in Table~\ref{tab:ablation_appendix} and the generated clusters from Table~\ref{tab:cluster_count_ablation}, we can see how the Dual-Prompt strategy is superior to single-prompt variants.

\begin{itemize}
  \item \textbf{\texttt{Relaxed Prompt} Variant:} This variant consistently generates a vastly larger number of clusters than the ground-truth (e.g., \textbf{1208} vs. 93 on ArxivS2S; \textbf{184} vs. 59 on Massive-I). This tendency to over-split the data results in poor semantic grouping and leads to the lowest average ARI of \textbf{32.5\%}.
  \item \textbf{\texttt{Strict Prompt} Variant:} In contrast, the variant is overly conservative, often producing a cluster count that is suboptimal for that dataset (e.g., only \textbf{41} clusters for MTOP-I, where the ground-truth is 102). While this consolidation can be beneficial, it often merges distinct topics, capping its average ARI at \textbf{38.1\%}.
  \item \textbf{\texttt{Dual-Prompt} Strategy:} The Dual-Prompt demonstrates a powerful adaptive capability. It navigates the trade-off between over-splitting and over-consolidating, producing a cluster count (e.g., \textbf{97} on Massive-I, \textbf{80} on MTOP-I) that better reflects the underlying data structure. This adaptive, dynamic control over granularity is the key reason it achieves the state-of-the-art average ARI of \textbf{45.4\%}, outperforming both single-prompt baselines by a significant margin.
\end{itemize}

\section{Detailed Robustness Analysis}
\label{app:robustness_detailed}

\subsection{Hyperparameter Sensitivity}
\label{app:hyperparam_sensitivity}

\begin{table*}[t!]
  \centering
  \setlength{\tabcolsep}{2.5pt}
  \resizebox{\textwidth}{!}{%
    \begin{tabular}{l ccc ccc ccc ccc ccc ccc ccc}
      \toprule
      \multirow{2}{*}{Offset} & \multicolumn{3}{c}{ArxivS2S} & \multicolumn{3}{c}{Massive-I} & \multicolumn{3}{c}{MTOP-I} & \multicolumn{3}{c}{Massive-D} & \multicolumn{3}{c}{FewNerd} & \multicolumn{3}{c}{FewRel} & \multicolumn{3}{c}{AVG}                                                                                                                                                                                                                                 \\
      \cmidrule(lr){2-4} \cmidrule(lr){5-7} \cmidrule(lr){8-10} \cmidrule(lr){11-13} \cmidrule(lr){14-16} \cmidrule(lr){17-19} \cmidrule(lr){20-22}
                              & ACC                          & NMI                           & ARI                        & ACC                           & NMI                         & ARI                        & ACC                     & NMI           & ARI           & ACC           & NMI           & ARI           & ACC           & NMI           & ARI           & ACC           & NMI           & ARI           & ACC           & NMI           & ARI           \\
      \midrule
      \textbf{0}              & \textbf{28.4}                & \textbf{57.4}                 & \textbf{16.3}              & \textbf{54.8}                 & \textbf{73.5}               & \textbf{47.9}              & \textbf{64.0}           & \textbf{77.5} & \textbf{68.9} & \textbf{57.6} & \textbf{67.7} & \textbf{53.8} & \textbf{59.3} & \textbf{63.3} & \textbf{53.1} & \textbf{43.2} & \textbf{63.6} & \textbf{32.7} & \textbf{51.2} & \textbf{67.2} & \textbf{45.4} \\
      \midrule
      -10                     & 27.4                         & 57.0                          & 15.9                       & 58.2                          & 72.9                        & 52.4                       & 59.7                    & 75.5          & 63.4          & 50.1          & 66.9          & 40.7          & 61.4          & 63.9          & 53.1          & 41.6          & 63.7          & 30.5          & 49.7          & 66.6          & 42.6          \\
      +10                     & 28.0                         & 58.6                          & 16.6                       & 53.8                          & 72.6                        & 45.9                       & 62.7                    & 76.2          & 65.9          & 55.6          & 68.2          & 50.6          & 56.6          & 61.0          & 49.4          & 39.6          & 60.8          & 27.3          & 49.4          & 66.2          & 42.6          \\
      +50                     & 25.0                         & 56.7                          & 14.2                       & 55.0                          & 74.6                        & 48.8                       & 61.2                    & 76.1          & 66.2          & 53.2          & 67.1          & 48.5          & 56.6          & 62.8          & 49.4          & 39.1          & 60.8          & 27.0          & 48.3          & 66.3          & 42.4          \\
      +100                    & 25.6                         & 57.3                          & 14.3                       & 53.0                          & 72.2                        & 44.4                       & 62.8                    & 75.7          & 65.7          & 49.4          & 68.3          & 42.1          & 54.2          & 59.9          & 49.7          & 40.7          & 61.4          & 29.4          & 47.6          & 65.8          & 40.9          \\
      +200                    & 27.7                         & 61.5                          & 16.7                       & 57.9                          & 74.4                        & 52.3                       & 63.7                    & 76.9          & 68.2          & 51.5          & 68.4          & 45.9          & 55.8          & 61.9          & 50.8          & 42.0          & 62.2          & 30.5          & 49.8          & 67.5          & 44.0          \\
      \bottomrule
    \end{tabular}%
  }
  \caption{Hyperparameter analysis of the dual-prompt transition threshold. We report ACC, NMI, and ARI (\%) across various switching offsets, with the default setting (offset 0) included for comparison.}
  \label{tab:hyperparam_appendix}
\end{table*}

Table~\ref{tab:hyperparam_appendix} details the full numerical results supporting the robustness claims in Section~\ref{ssec:hyperparam}.

\paragraph{Overall Performance Stability}
The average performance across all datasets demonstrates remarkable stability. The average ARI remains high across a wide spectrum of offsets, from an accelerated transition (offset -10, ARI \textbf{42.6\%}) to a significantly extended exploratory phase (offset +200, ARI \textbf{44.0\%}). The peak performance is achieved at the default offset of 0 (ARI \textbf{45.4\%}), but even extreme variations do not lead to a collapse in performance, underscoring the inherent self-correcting nature of the Dual-Prompt strategy.

\paragraph{Dataset-Specific Robustness}
The framework's demonstrated robustness is not merely a statistical artifact of averaging; rather, it is consistently evident at the individual dataset level.
\begin{itemize}
  \item On FewNerd, a semantically complex dataset, the ARI score proves to be exceptionally stable. It peaks at a high of \textbf{53.1\%} (offsets 0 and -10), while its lowest point remains a robust \textbf{49.4\%} (offset +10 and +50). This narrow range of fluctuation powerfully highlights the model's ability to achieve consistent clustering results, regardless of potential minor timing adjustments in the consolidation phase.
  \item On Massive-D, which exhibits more variance, performance still remains competitive. While the peak ARI of \textbf{53.8\%} is at the default offset, even an early transition (offset -10) yields a respectable ARI of \textbf{40.7\%}, and a late transition (offset +200) maintains an ARI of \textbf{45.9\%}.
  \item Notably, on some datasets like Massive-I, strategically shifting to an earlier transition (offset -10, ARI \textbf{52.4\%}) or a later one (offset +200, ARI \textbf{52.3\%}) can even outperform the default setting (ARI \textbf{47.9\%}), suggesting that while the default is a strong general-purpose choice, the framework is robust enough to accommodate diverse data distributions.
\end{itemize}

Table~\ref{tab:hyperparam_appendix} confirms that LLM-MemCluster is not dependent on precise hyperparameter tuning. Its strong performance across a wide range of offsets enables near-optimal results on diverse datasets without laborious optimization.

\subsection{Seed Robustness Under Randomized Dataset Execution Order}
\label{app:seed_robustness}

Table~\ref{tab:seed_robustness_gemini25_flash_lite} presents the stability of our framework under randomized dataset execution orders. We report the mean and standard deviation of ACC, NMI, and ARI across five random seeds using Gemini-2.5-Flash-Lite. These detailed statistics support the findings in Section~\ref{ssec:hyperparam}, confirming that the framework achieves highly consistent performance independent of the specific input sequence.

\newcommand{\meanstd}[2]{#1$\pm$#2}

\begin{table}[t!]
  \centering
  \resizebox{\columnwidth}{!}{
    \setlength{\tabcolsep}{4pt}
    \begin{tabular}{l ccc}
      \toprule
      Dataset   & ACC                   & NMI                   & ARI                   \\
      \midrule
      ArxivS2S  & \meanstd{30.13}{3.81} & \meanstd{53.52}{2.09} & \meanstd{15.82}{2.20} \\
      Massive-I & \meanstd{53.89}{2.36} & \meanstd{72.50}{2.17} & \meanstd{45.69}{2.85} \\
      MTOP-I    & \meanstd{64.46}{2.28} & \meanstd{79.50}{0.40} & \meanstd{67.14}{3.75} \\
      Massive-D & \meanstd{63.16}{2.77} & \meanstd{69.48}{1.72} & \meanstd{53.38}{4.05} \\
      FewNerd   & \meanstd{64.04}{2.47} & \meanstd{69.99}{2.30} & \meanstd{61.86}{2.06} \\
      FewRel    & \meanstd{30.81}{4.52} & \meanstd{54.92}{5.06} & \meanstd{19.34}{4.17} \\
      \bottomrule
    \end{tabular}
  }
  \caption{Robustness results using Gemini-2.5-Flash-Lite. Values represent mean $\pm$ std (\%) computed over 5 random permutations of the dataset.}
  \label{tab:seed_robustness_gemini25_flash_lite}
\end{table}

\section{Token Consumption and Cost}
\label{app:token_cost}

Table~\ref{tab:token_cost_gemini25_flash_lite} details the token consumption and estimated monetary cost incurred by our framework using Gemini-2.5-Flash-Lite. We report the aggregated input and output tokens for each dataset, alongside the total cost derived from public pricing at the time of experimentation (\$0.10 per 1M input tokens and \$0.40 per 1M output tokens).

While tokenization and pricing vary across models, a consistent structural feature of our framework is that operational costs are predominantly driven by inexpensive input tokens, whereas the volume of costly output tokens remains minimal.

\begin{table}[t!]
  \centering
  \resizebox{\columnwidth}{!}{
    \setlength{\tabcolsep}{2.5pt}
    \begin{tabular}{l rrr}
      \toprule
      Dataset        & Input Tokens        & Output Tokens    & Total Cost (USD) \\
      \midrule
      ArxivS2S       & 4,495,870           & 32,969           & 0.46             \\
      Massive-I      & 3,590,409           & 25,027           & 0.37             \\
      MTOP-I         & 4,537,051           & 33,653           & 0.47             \\
      Massive-D      & 3,231,711           & 31,226           & 0.34             \\
      FewNerd        & 4,821,283           & 33,393           & 0.50             \\
      FewRel         & 5,451,649           & 42,872           & 0.56             \\
      \midrule
      \textbf{Total} & \textbf{26,127,973} & \textbf{199,140} & \textbf{2.70}    \\
      \bottomrule
    \end{tabular}
  }
  \caption{Token consumption and estimated cost for running LLM-MemCluster with Gemini-2.5-Flash-Lite on each dataset. Input token price is \$0.10 per 1M tokens and output token price is \$0.40 per 1M tokens.}
  \label{tab:token_cost_gemini25_flash_lite}
\end{table}

\section{Algorithmic Implementation and Computational Complexity}
\label{app:algo_implementation}

The procedural implementation of our framework is detailed across two algorithms. Algorithm~\ref{alg:core_operation} describes the core, single-step clustering operation, which encapsulates the Dynamic Memory mechanism. Algorithm~\ref{alg:main_workflow} then presents the main workflow of LLM-MemCluster, illustrating how the core operation and the Dual-Prompt Granularity Control are integrated to process the entire dataset.

LLM-MemCluster processes a corpus of $N$ instances in a single, deterministic pass. For each instance, the primary computational costs stem from the LLM API call, $C_{LLM}$, and a potential retroactive update. A retroactive update incurs a cost of $O(i)$ at step $i$. While the theoretical worst-case complexity is $O(N^2)$, empirical evidence consistently shows that merge events are remarkably rare (averaging fewer than 2 per dataset in our experiments with Gemini-2.5-Flash-Lite). Consequently, the effective complexity remains near-linear, $O(N \cdot (C_{LLM} + C_{\text{update}}))$, avoiding the non-deterministic convergence behavior inherent to iterative algorithms like K-Means.

In contrast to contemporary methods like ClusterLLM, which employs a multi-stage pipeline to fine-tune a separate encoder, our framework is a unified, single-pass procedure. This design avoids the costly overhead of intermediate model training and multiple algorithmic phases.

\begin{algorithm}[t!]
  \caption{Core Clustering Operation}
  \label{alg:core_operation}
  \textbf{Input}: Text instance $x_i$; Memory of labels $\mathcal{M}_{\text{mem}}$; Assignments $\mathcal{A}$; Prompting $mode$ (Relaxed/Strict) \\
  \textbf{Output}: Updated $(\mathcal{M}_{\text{mem}}, \mathcal{A})$

  \begin{algorithmic}[1]
    \STATE $\mathcal{L}_{\text{seen}} \leftarrow \mathcal{M}_{\text{mem}}$
    \STATE $(l_i, s_i) \leftarrow \mathcal{F}_{\text{LLM}}(x_i, \mathcal{L}_{\text{seen}}, mode)$ \hfill {\footnotesize Eq.~(\ref{eq:core_call})}
    \STATE $\mathcal{A} \leftarrow \mathcal{A} \cup \{(x_i, l_i)\}$
    \IF{$l_i \notin \mathcal{L}_{\text{seen}}$}
    \STATE Add $l_i$ to memory $\mathcal{M}_{\text{mem}}$ \hfill {\footnotesize Eq.~(\ref{eq:reuse_create})}
    \ENDIF
    \IF{$s_i$ is not null}
    \STATE $\mathcal{L}_{old}, l_{new} \leftarrow$ Extract labels from $s_i$
    \STATE $\mathcal{M}_{\text{mem}} \leftarrow (\mathcal{M}_{\text{mem}}\backslash\mathcal{L}_{old})\cup\{l_{new}\}$ \hfill {\footnotesize Eq.~(\ref{eq:merge_refine})}
    \FOR{each $(x_j, l_j) \in \mathcal{A}$}
    \IF{$l_j \in \mathcal{L}_{old}$}
    \STATE $l_j \leftarrow l_{new}$ \hfill {\footnotesize Eq.~(\ref{eq:retroactive_update})}
    \ENDIF
    \ENDFOR
    \ENDIF
  \end{algorithmic}
\end{algorithm}

\begin{algorithm}[t!]
  \caption{LLM-MemCluster Workflow}
  \label{alg:main_workflow}
  \textbf{Input}: Unlabeled text corpus $\mathcal{D} = \{x_1, \dots, x_N\}$; LLM $\mathcal{F}_{\text{LLM}}$; Target K range $[K_{\min}, K_{\max}]$ \\
  \textbf{Output}: A partition of the corpus, $\mathcal{C}$.

  \begin{algorithmic}[1]
    \STATE Initialize memory $\mathcal{M}_{\text{mem}} \leftarrow \emptyset$ and assignments $\mathcal{A} \leftarrow \emptyset$
    \STATE \textit{\text{CoreOp} updates $\mathcal{M}_{\text{mem}}$ and $\mathcal{A}$ in-place.}
    \FOR{each text instance $x_i$ in $\mathcal{D}$}
    \IF{$|\mathcal{M}_{\text{mem}}| \ge K_{\max}$}
    \STATE $mode \leftarrow \text{Strict}$
    \ELSE
    \STATE $mode \leftarrow \text{Relaxed}$
    \ENDIF
    \STATE $\text{CoreOp}(x_i, \mathcal{M}_{\text{mem}}, \mathcal{A}, \text{mode})$
    \ENDFOR
    \STATE Generate the final partition $\mathcal{C}$ by grouping all instances in $\mathcal{A}$ by their assigned label
    \STATE \textbf{return} Final partition $\mathcal{C}$
  \end{algorithmic}
\end{algorithm}

\end{document}